\pdfoutput=1
\documentclass[11pt]{article}

\usepackage{acl}

\usepackage{times}
\usepackage{latexsym}

\usepackage[T1]{fontenc}
\usepackage[utf8]{inputenc}

\usepackage{microtype}

\usepackage{inconsolata}
\usepackage{booktabs}
\usepackage{graphicx}
\usepackage{float}
\usepackage{algorithm}
\usepackage{algorithmicx}  
\usepackage{algpseudocode}
\usepackage{tcolorbox}
\tcbuselibrary{skins, breakable, theorems}
\usepackage{CJKutf8}
\usepackage{multicol}
\usepackage{cuted}
\usepackage{hyperref}
\title{A Step Closer to Comprehensive Answers: Constrained Multi-Stage Question Decomposition with Large Language Models}

\author{
Hejing Cao\textsuperscript{\rm 1}\footnotemark[1],~
Zhenwei An\textsuperscript{\rm 1}\footnotemark[1],~
Jiazhan Feng\textsuperscript{\rm 1},~
Kun Xu\textsuperscript{\rm 2},~
Liwei Chen\textsuperscript{\rm 2},~
Dongyan Zhao\textsuperscript{\rm 1}\footnotemark[2]
\\
$^1$ Peking University
~~~~
$^2$ Kuaishou Technology \\
 \texttt {\{caohejing,anzhenwei,fengjiazhan,zhaody\}@pku.edu.cn}\\ 
}

\begin{document}
\maketitle

\renewcommand{\thefootnote}{\fnsymbol{footnote}}
\footnotetext[1]{Equal Contribution.}
\footnotetext[2]{Corresponding author.}

\begin{abstract}

While large language models exhibit remarkable performance in the Question Answering task, they are susceptible to hallucinations. Challenges arise when these models grapple with understanding multi-hop relations in complex questions or lack the necessary knowledge for a comprehensive response. To address this issue, we introduce the "Decompose-and-Query" framework (D\&Q). This framework guides the model to think and utilize external knowledge similar to ReAct, while also restricting its thinking to reliable information, effectively mitigating the risk of hallucinations. Experiments confirm the effectiveness of D\&Q: On our ChitChatQA dataset, D\&Q does not lose to ChatGPT in 67\% of cases; on the HotPotQA question-only setting, D\&Q achieved an F1 score of 59.6\%. Our code is available at \href{https://github.com/alkaidpku/DQ-ToolQA}{https://github.com/alkaidpku/DQ-ToolQA} .

% \hj{Comments here}
% \zw{Comments here}
% \jz{Comments here}
\end{abstract}

\section{Introduction}

Large language models (LLMs), such as ChatGPT \citep{chat-gpt}, GPT-4 \citep{openai_gpt-4_2023}, Bard \citep{bard}, have made significant inroads into people's work and daily life, substantially enhancing productivity. For LLMs, to produce high-quality response for a complex question relies on two essential capabilities: 1) Understanding multi-hop reasoning in complex questions; 2) Employing external knowledge.

On the one hand, many questions in our daily lives manifest as multi-hop questions. Individuals often pose intricate questions or instructions within a single sentence, lacking necessary descriptions. This necessitates LLMs to proficiently decompose complex questions into multiple sub-questions. One potential solution involves employing a sub-question decomposition approach, such as self-ask \citep{press_measuring_2023}, which decomposes the question into sub-questions, resolves them independently, and consolidates the answers. However, this approach relies on LLMs' ability to reasonably decompose questions and is typically more suitable for high-quality prompts, such as HotPotQA \citep{yang_hotpotqa_2018} and 2WikiMultiHopQA \citep{ho_constructing_2020}. Challenges may arise when dealing with low-quality or unclear prompts.

% \zw{On the other hand,  many methodologies incorporate thinking and external knowledge in LLMs to decompose complex question into sub-questions, ultimately leading to a comprehensive answer. However, answering these sub-questions may also bring hallucinations.}

On the other hand, large language models are prone to hallucinations~\citep{zhang2023siren} and do not have access to up-to-date knowledge~\citep{mallen_when_2023}. Integrating external tools (search engines, knowledge bases, APIs, and etc) with LLMs is a viable approach to address these challenges. Least-to-most~\citep{zhou_least--most_2023} decomposes complex question into sub-questions, ultimately leading to a comprehensive answer. However, answering these sub-questions may also bring hallucinations. ReAct~\citep{yao_react_2023} bridges the gap of reasoning and action, enabling thoughtful interactions with tools. It is noteworthy that the action space of ReAct is limited, and its effectiveness is relatively modest. \citet{shao_enhancing_2023-1} introduced Iter-RetGen, which iteratively follows a two-step process: retrieving relevant knowledge, and generating answer. The approach of Iter-RetGen bears similarities to Self-Refine~\citep{madaan2023selfrefine}. However, Iter-RetGen lacks autonomous termination capabilities.

To this, we have proposed a constrained question decomposition strategy that transforms question decomposition into a "Decompose-and-Query" depth-first search framework. Specifically, we construct a reliable question-answer base, which consists of high-quality question-answer pairs. The model is required to consult it to generate answers for search efficiency. In case of conflicts between the model’s knowledge and the reliable QA base, the base’s content takes precedence. 

We annotated solution trajectories for fine-tuning the model and subsequent predictions. The supervised fine-tuning phase does not involve tool invocation; the model processes static dialogue text, learning strategies for invoking tools. During the prediction phase, the model calls external tools to query a reliable QA base. If the LLM realizes during the search process that it has pursued an incorrect path, it can backtrack to previous steps and initiate a new search. Experiments confirm the effectiveness of D\&Q: On our ChitChatQA dataset, D\&Q does not lose to ChatGPT in 67\% of cases; on the HotPotQA question-only setting, D\&Q achieved an F1 score of 59.6\%.

Our contributions are the following: 

\begin{enumerate}
    \item We build and make public a dataset of real conversation scenarios: ChitChatQA.
    \item We design a framework for LLMs using external tools to perform constrained problem decomposition. Although simple, it can theoretically be extended to any number of external tools.
    % \item We draw the conclusion that supervised fine-tuning with a limited dataset is sufficient to enable tool invocation adaptation by Large Language Models (LLMs).
    \item We achieve an F1 score of 59.6\% on the question-only setting on HotPotQA dataset.
\end{enumerate}

\section{Related Works}

\paragraph{Reasoning.} The capacity for logical reasoning is a significant hallmark of intelligence in large language models. \citet{wei_chain--thought_2023} have proposed the Chain-of-Thought (CoT) as an effective means to enhance the reasoning abilities of large language models consistently and efficiently. However, incorrect reasoning paths often lead to incorrect reasoning results. To this, \citet{wang_self-consistency_2023} suggest self-consistency, which samples a diverse set of reasoning paths and selects the most consistent answer. Self-consistency, despite its higher decoding cost compared to CoT, still struggles with challenging problem-solving.
\citet{yao_tree_2023} introduce a novel approach by extending CoT into Tree-of-Thought (ToT). At each step, the model engages in self-evaluation and, when necessary, looks ahead or backtracks to enhance decision-making abilities. Similarly, Graph of Thoughts (GoT) framework models information generated by large language models as a graph, allowing for the combination of various LLM thoughts to yield collaborative results, extract insights from the entire thought network, and employ feedback loops to enhance thinking \citep{besta_graph_2023}. Cumulative reasoning \citep{zhang_cumulative_2023} employs proposers, verifiers, and reporters to store intermediate results, facilitating the accumulation of the reasoning process. So far, the reasoning capabilities of large language models have remained understudied.

\paragraph{Augmented Language Models.} Large language models are prone to generating hallucinations and do not have access to up-to-date knowledge. Large language models are likely to make factual errors with long-tail knowledge \citep{kandpal_large_2023}. As a consequence, more and more research is focusing on the use of external tools to enhance large language models. Retrieval-based language model is a classic augmented language model. The idea is to use discrete or dense retrievers to retrieve related context, and then merge the retrieved context into the input. It is possible to match certain capabilities of some of the largest LMs while having less parameters \citep{schick_toolformer_2023, qin_toolllm_2023}. Along with the research on CoT, many studies have integrated external tools into the chain of thought paradigm \citep{yao_react_2023, paranjape_art_2023, feng_knowledge_2023, press_measuring_2023, shao_enhancing_2023, trivedi_interleaving_2023, gou_critic_2023, gou_tora_2023}. Synergizing reasoning and invocation tools in LLMs is an underexplored potential research domain.

\begin{figure*}[ht]
    \centering
    \includegraphics[width=1.0\textwidth]{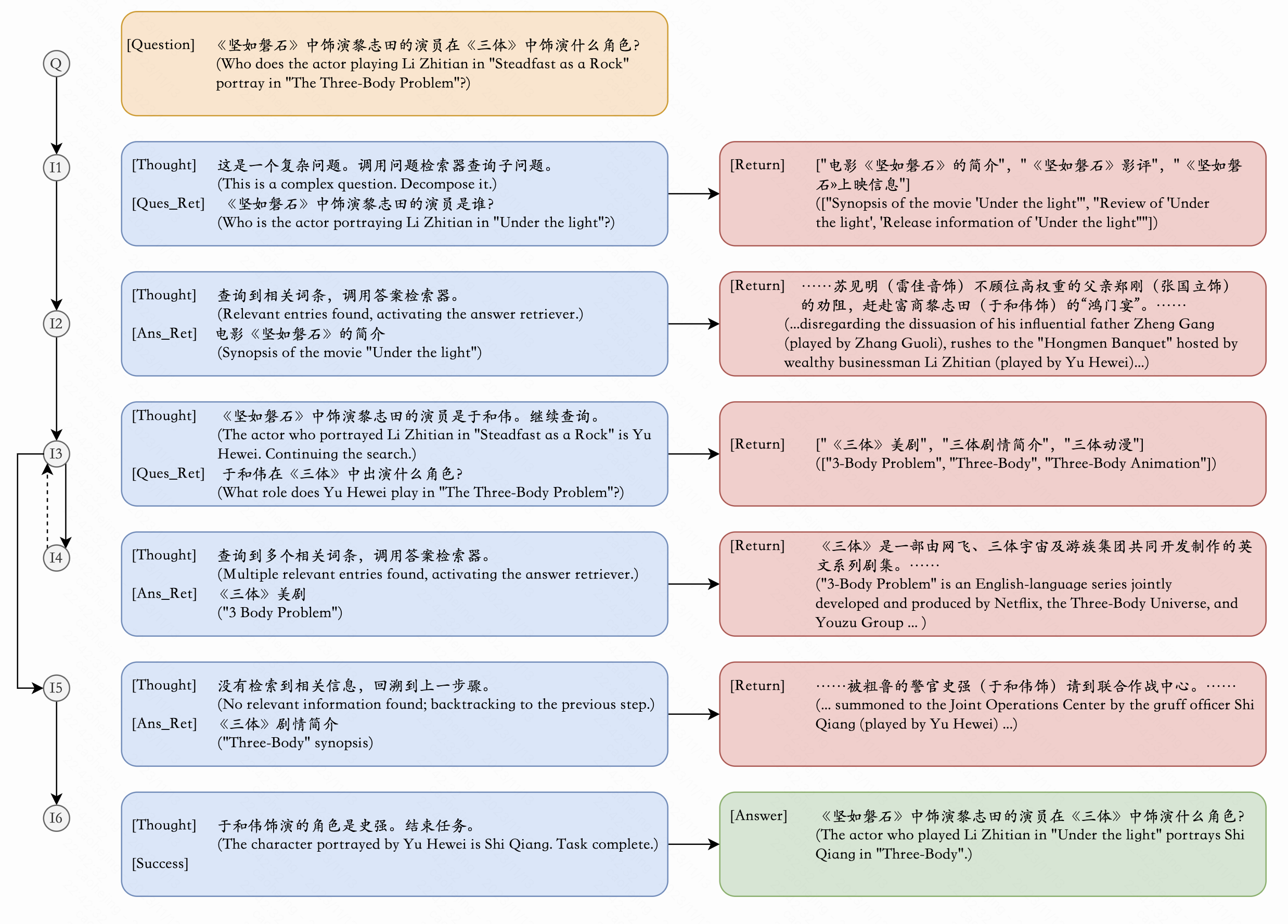}
    \caption{Example of question decomposition trajectory. Q: Question; I: Iteration. Dashed line means rollback. We use Ques\_Ret for QuestionRetriever and Ans\_Ret for AnswerRetriever for brevity.
    }
    \label{fig:example}
\end{figure*}

\section{Methodology}

This section presents our methodology. We construct ChitChatQA and the reliable QA base, annotate problem-solving trajectories and train a model to invoke tools.

\subsection{ChitChatQA dataset} \label{sec:3.1}

We introduce a novel dataset, ChitChatQA, to assess the capabilities of large language models in handling complex and diverse questions. Specifically, we collected single-turn dialogue data from real scenarios in the KwaiTalk application. The dataset comprises authentic questions posed by users and responses generated by large language models. We denote the original question-answer data as $S_0 = {(q_i, a_i)}_{N_0}$.

We performed data cleaning on the collected question-answer dataset to eliminate noise. Specifically, we initially utilized a grammatical correction model to filter out questions with low scores, followed by an intent recognition model to screen for questions with ambiguous intent. During the data cleaning process, we applied a lower intent recognition threshold to retain data with unclear intents. The detailed procedure for data cleaning can be found in Algorithm \ref{alg:construction}.

\begin{algorithm}[htb]
\caption{Reliable QA Database Creation}
\label{alg:construction}

\begin{algorithmic}[1]
    \Require $S_0$, $\epsilon_1$, $\epsilon_2$, GEC, Intent
    \Ensure $S$
\item $S$ = new Set()
\For{$q_i$ in $S_0$}
    \If{GEC$(q_i) > \epsilon_1$ and Intent$(q_i) > \epsilon_2$} 
        \State $S$.Add($q_i$)
    \EndIf
\EndFor
\State \textbf{return} $S$
\end{algorithmic}
\end{algorithm}

\begin{figure*}[ht]
    \centering
    \includegraphics[width=1.0\textwidth]{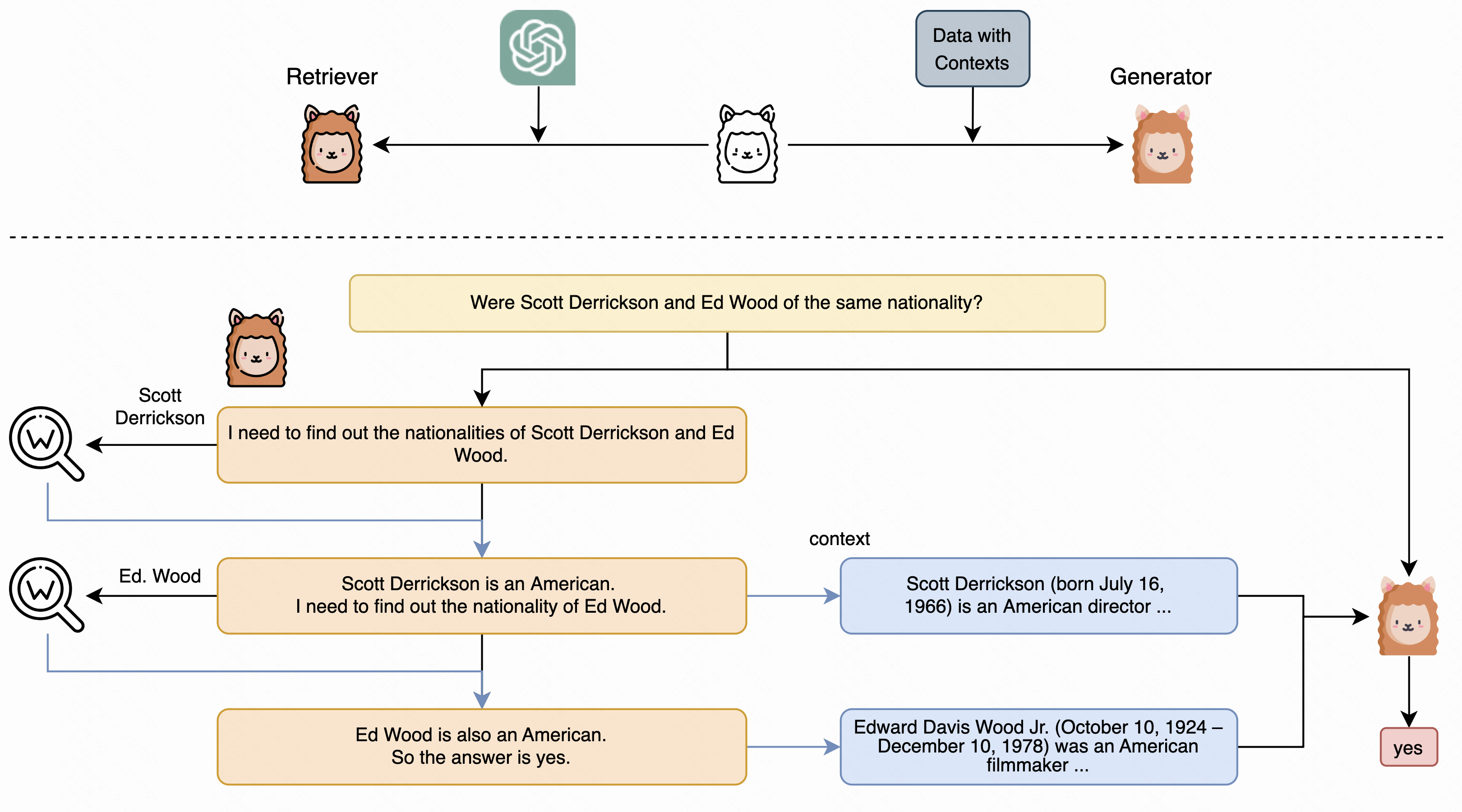}
    \caption{Model architecture of HotPotQA.}
    \label{fig:hotpotqa}
\end{figure*}

\subsection{Reliable QA Base} \label{sec:3.2}

We constructed a reliable question-answer base, which consists of high-quality question-answer pairs. Similar to Constitutional AI~\citep{bai_constitutional_2022}, all question-answer pairs are considered beyond doubt. The model is required to consult the database to generate answers. In case of conflicts between the model's knowledge and the reliable QA base, the base's content takes precedence. We believe that this makes the model less prone to hallucinations and more robust.

In dialogue system, users may inquire about similar questions, such as "please provide me with a Python code for Bubble Sort" or "tell me the names of the Eight Great Masters of the Tang and Song dynasties". These generic questions are frequent. Preliminary analysis shows that these questions exhibit high quality. Therefore, we selected the top 50,000 questions based on their frequency as benchmark questions for the reliable QA base.

Due to the fact that the pretraining data for large language models comprises noise, the outputs for the same question may vary. We categorized the questions into several types. Objective factual questions demand precise and unique responses. For these questions, we employ a majority voting approach to determine the response. Subjective opinion questions pertain to opinions or perspectives on certain matters, and for such questions, we use an opinion aggregation method to determine the response. Consequently, we designed an answer aggregation model to consolidate diverse viewpoints.

The answer aggregation model is essentially a large language model through supervised fine-tuning. We manually annotated 5,000 answer aggregation instances. Leveraging the LLaMA2-13b model as the base, we fine-tuned the model on the aforementioned data to create the answer aggregation model. Subsequently, we performed answer aggregation on all the question-answer pairs in the reliable question-answer base.

For examples of answer aggregation, please refer to Appendix \ref{sec:appendixA}.

\subsection{Trajectory Annotation} \label{sec:3.3}

To train the model to invoke external tools, we need gather a sufficient amount of data for supervised fine-tuning. Both ChatGPT and GPT-4 exhibit excellent performance in data annotation ~\citep{ubani2023zeroshotdataaug}. Based on preliminary experiments, the annotation capabilities of ChatGPT are significantly inferior to those of GPT-4. The annotation framework is illustrated in Figure \ref{fig:example}.

\subsection{Adaptive Constraint Question Decomposition} \label{sec:3.4}

We employ imitation learning using data that involves multiple iterations of calling an external retriever for fine-tuning large language models. We use LLaMA2-13b as base model. The training is conducted in a multi-turn dialogue format, wherein, in each training round, only the output of that round's dialogue is retained, and the rest is masked. For more details, please refer to the Experimental Section.

\section{Experiments}

\begin{table*}[htb]
\centering
\begin{tabular}{@{}lllll@{}}
\toprule
Method                                                     & Generation Model & Model Size & EM   & F1   \\ \midrule
Standard                                                   & PaLM             & 540B       & 21.9 & 36.8 \\
CoT \citep{wei2023chainofthought}                         & PaLM             & 540B       & 30.0 & 44.1 \\
ReAct \citep{yao_react_2023}            & PaLM             & 540B       & 24.9 & 44.7 \\
Self-Ask \citep{press_measuring_2023}   & InstructGPT      & 175B       & 36.8 & 55.2 \\
IRCoT \citep{trivedi_interleaving_2023} & Code-Davinci-002 & 175B       & -    & 60.7 \\
D\&Q                                                       & LLaMA2           & 13B        & 39.0 & 59.6 \\ \bottomrule
\end{tabular}
\caption{Results on HotPotQA dataset. EM: Exact Match.}
\label{tab:hotpotqa_result}
\end{table*}

Our primary experiments consist of two parts. To evaluate question decomposition capability, experiments were conducted on the ChitChatQA dataset. To assess the effectiveness of our approach in answering multi-hop questions, experiments were conducted using the academic dataset HotPotQA.

\subsection{ChitChatQA}

We utilized a set of 200 training instances and conducted supervised fine-tuning in a multi-turn dialogue format on LLaMA2-13b. The supervised fine-tuning phase does not involve tool invocation; the model processes static dialogue text, learning strategies for invoking tools. During the prediction phase, the model calls external tools to query the reliable QA base. We defined three tools: \texttt{[QuestionRetriever]} for retrieving related questions from the QA base, \texttt{[AnswerRetriever]} for retrieve answer to a question, and \texttt{[Finish]} indicating task termination.

ChitChatQA's data stems from real dialogue interactions with the Kuaishou dialogue model, with a test set comprising 100 instances. We employed the Kuaishou dialogue model's output as a baseline, noted as "KwaiDialog", and compared it with ChatGPT's output. Model evaluation was conducted through human assessments labeled as win/tie/lose, as shown in Table 1.

The results indicate a significant enhancement in the response capability of the Base model with the D\&Q approach. Even after complex prompt fine-tuning and reinforcement learning for KwaiDialog, our method still achieves victory in 37\% of test cases (not losing in 73\%). Compared to ChatGPT, D\&Q maintains a competitive position, not losing in 62\% of cases. 
% Detailed sample analyses are provided in Appendix A.

\begin{table}[ht]
\centering
\begin{tabular}{@{}cccc@{}}
\toprule
                   & win  & tie  & lose \\ \midrule
D\&Q vs KwaiDialog & 37\% & 36\% & 27\% \\
D\&Q vs ChatGPT    & 24\% & 38\% & 38\% \\ \bottomrule
\end{tabular}
\caption{Human evaluation on ChitChatQA}
\end{table}

\subsection{HotPotQA}

\subsubsection{Baselines}

\paragraph{ReAct}    Utilizes Search and Lookup tools iteratively to provide external information for large language models \citep{yao_react_2023}.

\paragraph{Self-Ask}    Requires the model to ask itself and use the answers by the search engine to enhance the model \citep{press_measuring_2023}.

\paragraph{IRCoT}    Decomposes questions through Chain-of-Thought, retrieves results using a retriever, then concatenates the question and reference context to a more robust language model (GPT-3).

\subsubsection{Implementation Details}

We conducted experiments on the HotPotQA dataset. Similar to most baseline models, our configuration was question-only: providing only the question without accompanying references, allowing the model to autonomously invoke external tools for information retrieval. We defined three tools: \texttt{[ArticleRetriever]} for querying entries in Wikipedia, \texttt{[PageRetriever]} for retrieving pages corresponding to the entries, and \texttt{[Finish]} representing task termination. The specific implementations of these tools were achieved through calls to the MediaWiki\footnote{https://www.mediawiki.org/wiki/API:Search/} API. We utilized 200 training instances and performed supervised fine-tuning in a multi-turn dialogue format on LLaMA2-13b.

\subsubsection{Results}

Table \ref{tab:hotpotqa_result} shows the D\&Q result of HotPotQA. We can draw the following conclusions:

\textbf{Backtracking enhances the model's retrieval capability.} Although ReAct supports acquiring reference information through observation, this approach is, in practice, greedy. Specifically, ReAct does not explicitly backtrack but tends toward random walks. In contrast, D\&P stores intermediate states, allowing it to retreat to the previous state in case of retrieval failure, significantly enhancing retrieval capability.

\textbf{Tool invocation can compensate for parameter limitations.} In comparison to IRCoT's GPT-3 (175B), our generative model is LLaMA-13B. Through supervised fine-tuning, D\&Q can achieve results comparable to IRCoT. This indicates that smaller models, through supervised fine-tuning to learn tool invocation, can achieve performance comparable to larger models.

\section{Ablation Study}

\subsection{Retrieval Ability}

\begin{table}[ht]
\centering
\begin{tabular}{@{}lcc@{}}
\toprule
Method        & Average Contexts & Recall \\ \midrule
direct prompt & 26.7             & 52.3   \\
D\&Q          & 50.0             & 68.8   \\ \bottomrule
\end{tabular}
\caption{Recall of retrieval.}
\label{tab:ablation}
\end{table}

We limit the retriever's calls to a maximum of 10, with each call returning up to 5 relevant entries. Therefore, the model can return a maximum of 50 entries when generating responses. We calculate the recall by retrieving 50 entries using the initial question as a baseline. We tested the recall of D\&Q's retrieval ability and compared it with the baseline, resulting in an increase from 52.3\% to 68.8\%. This indicates that considering context contributes to more accurate tool invocation.

\section{Conclusion}

In conclusion, we propose the "Decompose-and-Query" framework (D\&Q) to address the susceptibility of large language models to hallucinations by guiding them to utilize external knowledge within a constrained and reliable context. Through experiments on the ChitChatQA dataset and the HotPotQA question-only setting, D\&Q demonstrated competitive performance against existing models. Our contributions include the introduction of the ChitChatQA dataset, a novel framework for constrained problem decomposition, and the achievement of a notable F1 score on HotPotQA. These findings underscore the potential of D\&Q in enhancing the robustness and performance of large language models in question-answering tasks.

\bibliography{custom}

\begin{thebibliography}{29}
\expandafter\ifx\csname natexlab\endcsname\relax\def\natexlab#1{#1}\fi

\bibitem[{Bai et~al.(2022)Bai, Kadavath, Kundu, Askell, Kernion, Jones, Chen, Goldie, Mirhoseini, McKinnon, Chen, Olsson, Olah, Hernandez, Drain, Ganguli, Li, Tran-Johnson, Perez, Kerr, Mueller, Ladish, Landau, Ndousse, Lukosuite, Lovitt, Sellitto, Elhage, Schiefer, Mercado, DasSarma, Lasenby, Larson, Ringer, Johnston, Kravec, Showk, Fort, Lanham, Telleen-Lawton, Conerly, Henighan, Hume, Bowman, Hatfield-Dodds, Mann, Amodei, Joseph, McCandlish, Brown, and Kaplan}]{bai_constitutional_2022}
Yuntao Bai, Saurav Kadavath, Sandipan Kundu, Amanda Askell, Jackson Kernion, Andy Jones, Anna Chen, Anna Goldie, Azalia Mirhoseini, Cameron McKinnon, Carol Chen, Catherine Olsson, Christopher Olah, Danny Hernandez, Dawn Drain, Deep Ganguli, Dustin Li, Eli Tran-Johnson, Ethan Perez, Jamie Kerr, Jared Mueller, Jeffrey Ladish, Joshua Landau, Kamal Ndousse, Kamile Lukosuite, Liane Lovitt, Michael Sellitto, Nelson Elhage, Nicholas Schiefer, Noemi Mercado, Nova DasSarma, Robert Lasenby, Robin Larson, Sam Ringer, Scott Johnston, Shauna Kravec, Sheer~El Showk, Stanislav Fort, Tamera Lanham, Timothy Telleen-Lawton, Tom Conerly, Tom Henighan, Tristan Hume, Samuel~R. Bowman, Zac Hatfield-Dodds, Ben Mann, Dario Amodei, Nicholas Joseph, Sam McCandlish, Tom Brown, and Jared Kaplan. 2022.
\newblock \href {https://arxiv.org/abs/2212.08073} {Constitutional {AI}: {Harmlessness} from {AI} {Feedback}}.

\bibitem[{Besta et~al.(2023)Besta, Blach, Kubicek, Gerstenberger, Gianinazzi, Gajda, Lehmann, Podstawski, Niewiadomski, Nyczyk, and Hoefler}]{besta_graph_2023}
Maciej Besta, Nils Blach, Ales Kubicek, Robert Gerstenberger, Lukas Gianinazzi, Joanna Gajda, Tomasz Lehmann, Michal Podstawski, Hubert Niewiadomski, Piotr Nyczyk, and Torsten Hoefler. 2023.
\newblock \href {https://arxiv.org/abs/2308.09687} {Graph of {Thoughts}: {Solving} {Elaborate} {Problems} with {Large} {Language} {Models}}.

\bibitem[{Feng et~al.(2023)Feng, Tao, Geng, Shen, Xu, Long, Zhao, and Jiang}]{feng_knowledge_2023}
Jiazhan Feng, Chongyang Tao, Xiubo Geng, Tao Shen, Can Xu, Guodong Long, Dongyan Zhao, and Daxin Jiang. 2023.
\newblock \href {https://arxiv.org/abs/2305.07402} {Knowledge {Refinement} via {Interaction} {Between} {Search} {Engines} and {Large} {Language} {Models}}.

\bibitem[{Google(2023)}]{bard}
Google. 2023.
\newblock {Google bard}.
\newblock \url{https://bard.google.com/}.
\newblock URL https://bard.google.com/.

\bibitem[{Gou et~al.(2023{\natexlab{a}})Gou, Shao, Gong, Shen, Yang, Duan, and Chen}]{gou_critic_2023}
Zhibin Gou, Zhihong Shao, Yeyun Gong, Yelong Shen, Yujiu Yang, Nan Duan, and Weizhu Chen. 2023{\natexlab{a}}.
\newblock \href {https://arxiv.org/abs/2305.11738} {{CRITIC}: {Large} {Language} {Models} {Can} {Self}-{Correct} with {Tool}-{Interactive} {Critiquing}}.

\bibitem[{Gou et~al.(2023{\natexlab{b}})Gou, Shao, Gong, Shen, Yang, Huang, Duan, and Chen}]{gou_tora_2023}
Zhibin Gou, Zhihong Shao, Yeyun Gong, Yelong Shen, Yujiu Yang, Minlie Huang, Nan Duan, and Weizhu Chen. 2023{\natexlab{b}}.
\newblock \href {https://arxiv.org/abs/2309.17452} {{ToRA}: {A} {Tool}-{Integrated} {Reasoning} {Agent} for {Mathematical} {Problem} {Solving}}.

\bibitem[{Ho et~al.(2020)Ho, Duong~Nguyen, Sugawara, and Aizawa}]{ho_constructing_2020}
Xanh Ho, Anh-Khoa Duong~Nguyen, Saku Sugawara, and Akiko Aizawa. 2020.
\newblock \href {https://doi.org/10.18653/v1/2020.coling-main.580} {Constructing a multi-hop {QA} dataset for comprehensive evaluation of reasoning steps}.
\newblock In \emph{Proceedings of the 28th International Conference on Computational Linguistics}, pages 6609--6625, Barcelona, Spain (Online). International Committee on Computational Linguistics.

\bibitem[{Kandpal et~al.(2022)Kandpal, Deng, Roberts, Wallace, and Raffel}]{kandpal_large_2023}
Nikhil Kandpal, Haikang Deng, Adam Roberts, Eric Wallace, and Colin Raffel. 2022.
\newblock \href {https://arxiv.org/abs/2211.08411} {Large {Language} {Models} {Struggle} to {Learn} {Long}-{Tail} {Knowledge}}.

\bibitem[{Madaan et~al.(2023)Madaan, Tandon, Gupta, Hallinan, Gao, Wiegreffe, Alon, Dziri, Prabhumoye, Yang, Gupta, Majumder, Hermann, Welleck, Yazdanbakhsh, and Clark}]{madaan2023selfrefine}
Aman Madaan, Niket Tandon, Prakhar Gupta, Skyler Hallinan, Luyu Gao, Sarah Wiegreffe, Uri Alon, Nouha Dziri, Shrimai Prabhumoye, Yiming Yang, Shashank Gupta, Bodhisattwa~Prasad Majumder, Katherine Hermann, Sean Welleck, Amir Yazdanbakhsh, and Peter Clark. 2023.
\newblock \href {http://arxiv.org/abs/2303.17651} {Self-refine: Iterative refinement with self-feedback}.

\bibitem[{Mallen et~al.(2023)Mallen, Asai, Zhong, Das, Khashabi, and Hajishirzi}]{mallen_when_2023}
Alex Mallen, Akari Asai, Victor Zhong, Rajarshi Das, Daniel Khashabi, and Hannaneh Hajishirzi. 2023.
\newblock \href {https://doi.org/10.48550/arXiv.2212.10511} {When {Not} to {Trust} {Language} {Models}: {Investigating} {Effectiveness} of {Parametric} and {Non}-{Parametric} {Memories}}.
\newblock ArXiv:2212.10511 [cs].

\bibitem[{OpenAI(2022)}]{chat-gpt}
OpenAI. 2022.
\newblock {Chatgpt: Optimizing language models for dialogue}.
\newblock \url{https://openai.com/blog/chatgpt/}.
\newblock URL https://openai.com/blog/chatgpt/.

\bibitem[{OpenAI(2023)}]{openai_gpt-4_2023}
OpenAI. 2023.
\newblock \href {https://arxiv.org/abs/2303.08774} {{GPT}-4 {Technical} {Report}}.

\bibitem[{Paranjape et~al.(2023)Paranjape, Lundberg, Singh, Hajishirzi, Zettlemoyer, and Ribeiro}]{paranjape_art_2023}
Bhargavi Paranjape, Scott Lundberg, Sameer Singh, Hannaneh Hajishirzi, Luke Zettlemoyer, and Marco~Tulio Ribeiro. 2023.
\newblock \href {https://arxiv.org/abs/2303.09014} {{ART}: {Automatic} multi-step reasoning and tool-use for large language models}.

\bibitem[{Press et~al.(2022)Press, Zhang, Min, Schmidt, Smith, and Lewis}]{press_measuring_2023}
Ofir Press, Muru Zhang, Sewon Min, Ludwig Schmidt, Noah~A. Smith, and Mike Lewis. 2022.
\newblock \href {https://arxiv.org/abs/2210.03350} {Measuring and {Narrowing} the {Compositionality} {Gap} in {Language} {Models}}.

\bibitem[{Qin et~al.(2023)Qin, Liang, Ye, Zhu, Yan, Lu, Lin, Cong, Tang, Qian, Zhao, Tian, Xie, Zhou, Gerstein, Li, Liu, and Sun}]{qin_toolllm_2023}
Yujia Qin, Shihao Liang, Yining Ye, Kunlun Zhu, Lan Yan, Yaxi Lu, Yankai Lin, Xin Cong, Xiangru Tang, Bill Qian, Sihan Zhao, Runchu Tian, Ruobing Xie, Jie Zhou, Mark Gerstein, Dahai Li, Zhiyuan Liu, and Maosong Sun. 2023.
\newblock \href {https://arxiv.org/abs/2307.16789} {{ToolLLM}: {Facilitating} {Large} {Language} {Models} to {Master} 16000+ {Real}-world {APIs}}.

\bibitem[{Schick et~al.(2023)Schick, Dwivedi-Yu, Dessì, Raileanu, Lomeli, Zettlemoyer, Cancedda, and Scialom}]{schick_toolformer_2023}
Timo Schick, Jane Dwivedi-Yu, Roberto Dessì, Roberta Raileanu, Maria Lomeli, Luke Zettlemoyer, Nicola Cancedda, and Thomas Scialom. 2023.
\newblock \href {https://arxiv.org/abs/2302.04761} {Toolformer: {Language} {Models} {Can} {Teach} {Themselves} to {Use} {Tools}}.

\bibitem[{Shao et~al.(2023{\natexlab{a}})Shao, Gong, Shen, Huang, Duan, and Chen}]{shao_enhancing_2023-1}
Zhihong Shao, Yeyun Gong, Yelong Shen, Minlie Huang, Nan Duan, and Weizhu Chen. 2023{\natexlab{a}}.
\newblock \href {https://arxiv.org/abs/2305.15294} {Enhancing {Retrieval}-{Augmented} {Large} {Language} {Models} with {Iterative} {Retrieval}-{Generation} {Synergy}}.

\bibitem[{Shao et~al.(2023{\natexlab{b}})Shao, Gong, Shen, Huang, Duan, and Chen}]{shao_enhancing_2023}
Zhihong Shao, Yeyun Gong, Yelong Shen, Minlie Huang, Nan Duan, and Weizhu Chen. 2023{\natexlab{b}}.
\newblock \href {https://arxiv.org/abs/2305.15294} {Enhancing {Retrieval}-{Augmented} {Large} {Language} {Models} with {Iterative} {Retrieval}-{Generation} {Synergy}}.

\bibitem[{Trivedi et~al.(2022)Trivedi, Balasubramanian, Khot, and Sabharwal}]{trivedi_interleaving_2023}
Harsh Trivedi, Niranjan Balasubramanian, Tushar Khot, and Ashish Sabharwal. 2022.
\newblock \href {https://arxiv.org/abs/2212.10509} {Interleaving {Retrieval} with {Chain}-of-{Thought} {Reasoning} for {Knowledge}-{Intensive} {Multi}-{Step} {Questions}}.

\bibitem[{Ubani et~al.(2023)Ubani, Polat, and Nielsen}]{ubani2023zeroshotdataaug}
Solomon Ubani, Suleyman~Olcay Polat, and Rodney Nielsen. 2023.
\newblock \href {http://arxiv.org/abs/2304.14334} {Zeroshotdataaug: Generating and augmenting training data with chatgpt}.

\bibitem[{Wang et~al.(2022)Wang, Wei, Schuurmans, Le, Chi, Narang, Chowdhery, and Zhou}]{wang_self-consistency_2023}
Xuezhi Wang, Jason Wei, Dale Schuurmans, Quoc Le, Ed~Chi, Sharan Narang, Aakanksha Chowdhery, and Denny Zhou. 2022.
\newblock \href {https://arxiv.org/abs/2203.11171} {Self-{Consistency} {Improves} {Chain} of {Thought} {Reasoning} in {Language} {Models}}.

\bibitem[{Wei et~al.(2022)Wei, Wang, Schuurmans, Bosma, Ichter, Xia, Chi, Le, and Zhou}]{wei_chain--thought_2023}
Jason Wei, Xuezhi Wang, Dale Schuurmans, Maarten Bosma, Brian Ichter, Fei Xia, Ed~Chi, Quoc Le, and Denny Zhou. 2022.
\newblock \href {https://arxiv.org/abs/2201.11903} {Chain-of-{Thought} {Prompting} {Elicits} {Reasoning} in {Large} {Language} {Models}}.

\bibitem[{Wei et~al.(2023)Wei, Wang, Schuurmans, Bosma, Ichter, Xia, Chi, Le, and Zhou}]{wei2023chainofthought}
Jason Wei, Xuezhi Wang, Dale Schuurmans, Maarten Bosma, Brian Ichter, Fei Xia, Ed~Chi, Quoc Le, and Denny Zhou. 2023.
\newblock \href {http://arxiv.org/abs/2201.11903} {Chain-of-thought prompting elicits reasoning in large language models}.

\bibitem[{Yang et~al.(2018)Yang, Qi, Zhang, Bengio, Cohen, Salakhutdinov, and Manning}]{yang_hotpotqa_2018}
Zhilin Yang, Peng Qi, Saizheng Zhang, Yoshua Bengio, William Cohen, Ruslan Salakhutdinov, and Christopher~D. Manning. 2018.
\newblock \href {https://doi.org/10.18653/v1/D18-1259} {{H}otpot{QA}: A dataset for diverse, explainable multi-hop question answering}.
\newblock In \emph{Proceedings of the 2018 Conference on Empirical Methods in Natural Language Processing}, pages 2369--2380, Brussels, Belgium. Association for Computational Linguistics.

\bibitem[{Yao et~al.(2023)Yao, Yu, Zhao, Shafran, Griffiths, Cao, and Narasimhan}]{yao_tree_2023}
Shunyu Yao, Dian Yu, Jeffrey Zhao, Izhak Shafran, Thomas~L. Griffiths, Yuan Cao, and Karthik Narasimhan. 2023.
\newblock \href {https://arxiv.org/abs/2305.10601} {Tree of {Thoughts}: {Deliberate} {Problem} {Solving} with {Large} {Language} {Models}}.

\bibitem[{Yao et~al.(2022)Yao, Zhao, Yu, Du, Shafran, Narasimhan, and Cao}]{yao_react_2023}
Shunyu Yao, Jeffrey Zhao, Dian Yu, Nan Du, Izhak Shafran, Karthik Narasimhan, and Yuan Cao. 2022.
\newblock \href {https://arxiv.org/abs/2210.03629} {{ReAct}: {Synergizing} {Reasoning} and {Acting} in {Language} {Models}}.

\bibitem[{Zhang et~al.(2023{\natexlab{a}})Zhang, Yang, Yuan, and Yao}]{zhang_cumulative_2023}
Yifan Zhang, Jingqin Yang, Yang Yuan, and Andrew Chi-Chih Yao. 2023{\natexlab{a}}.
\newblock \href {https://arxiv.org/abs/2308.04371} {Cumulative {Reasoning} with {Large} {Language} {Models}}.

\bibitem[{Zhang et~al.(2023{\natexlab{b}})Zhang, Li, Cui, Cai, Liu, Fu, Huang, Zhao, Zhang, Chen et~al.}]{zhang2023siren}
Yue Zhang, Yafu Li, Leyang Cui, Deng Cai, Lemao Liu, Tingchen Fu, Xinting Huang, Enbo Zhao, Yu~Zhang, Yulong Chen, et~al. 2023{\natexlab{b}}.
\newblock Siren's song in the ai ocean: A survey on hallucination in large language models.
\newblock \emph{arXiv preprint arXiv:2309.01219}.

\bibitem[{Zhou et~al.(2022)Zhou, Schärli, Hou, Wei, Scales, Wang, Schuurmans, Cui, Bousquet, Le, and Chi}]{zhou_least--most_2023}
Denny Zhou, Nathanael Schärli, Le~Hou, Jason Wei, Nathan Scales, Xuezhi Wang, Dale Schuurmans, Claire Cui, Olivier Bousquet, Quoc Le, and Ed~Chi. 2022.
\newblock \href {https://arxiv.org/abs/2205.10625} {Least-to-{Most} {Prompting} {Enables} {Complex} {Reasoning} in {Large} {Language} {Models}}.

\end{thebibliography}

\clearpage
\appendix
\newpage
\section{Answer Aggregation Data}
\label{sec:appendixA}

\begin{tcolorbox}[width=\textwidth]
\begin{CJK}{UTF8}{gbsn}
% \tiny
% \scriptsize
% \footnotesize
% \small
% \normalsize
% \large
% \Large
% \LARGE
% \huge
% \Huge
\scriptsize
问题：七夕节是干什么的？

答案1：七夕节，又称为“牛郎织女节”或“中国情人节”，是中国的一个传统节日。它起源于古代的牛郎织女神话，相传织女是王母娘娘的第七个女儿，与牛郎相爱，却被银河分隔，两人只能每年七夕这天才能相会一次。这个节日主要用于祈求姻缘、团圆和幸福。在七夕节这一天，人们会举行各种庆祝活动，如赏花灯、游灯会、拜织女、吃巧果等。现在，七夕节也成为了年轻人表达爱意和浪漫氛围的一个节日。

答案2：七夕节，又称为“牛郎织女节”或“中国情人节”，是中国的一个传统节日。它起源于古代的星宿崇拜，传说牛郎（天河星）和织女（织女星）在天河相会。这个故事表现了人们对美好爱情的向往和祈愿。七夕节通常在农历七月七日这一天庆祝，人们在这一天会举行各种活动，如祈求姻缘、情侣相会、送礼物等。其中最著名的活动是“拜七姐”，即向织女祈求智慧和巧艺。此外，还有放风筝、观星、赏花灯等传统习俗。

答案3：七夕节，又称乞巧节、女儿节，是中国传统的节日，起源于古代对天文星宿的崇拜。七夕节在农历七月初七，传说这一天是牛郎织女鹊桥相会的日子。在这一天，许多人会通过各种方式祈求自己的爱情美满，如织女星的祈求、乞巧、拜织女、吃巧果等。同时，七夕节也是中国的情人节，许多情侣会在这一天表达爱意，送礼物、吃烛光晚餐等。

答案4：七夕节是中国的传统节日，人们会进行许多习俗活动，如折桃枝、拜织女、揉木槿、吃巧果、洗梳具、染指甲、贺牛生日、拼七巧板、穿针乞巧、种生求子、供奉磨喝乐等。此外，七夕节也是中国的情人节，人们会进行乞巧、拜仙、穿针等仪式，祈求心灵手巧、获得美满姻缘。

答案5：七夕节是中国传统的情人节，也被称为“女儿节”或“乔奇日”，类似于中国情人节中的2月14日情人节。每年的七夕是牛郎织女相会的日子，这是一个美丽而永恒的爱情故事。

答案6：七夕节是中国传统节日中最具浪漫色彩的一个节日，主要参与者是少女，节日活动以乞巧为主，故被称为“乞巧节”、“少女节”、“女儿节”。在这一天晚上，妇女们穿针乞巧，祈祷福禄寿，礼拜七姐，仪式虔诚而隆重，陈列花果女红，各式家具、用具都精美小巧、惹人喜爱。

答案7：七夕节是中国的传统节日，人们会举行各种仪式和活动，如穿针乞巧、喜蛛应巧、投针验巧、种生求子、供奉“磨喝乐”、拜织女、拜魁星、晒书、晒衣、贺牛生日、吃巧果等等。此外，七夕节还有各种小玩偶摆在家里，穿针比赛是最早开始流行的，通过穿针的方式乞巧，穿的最快者就是手最巧的。

答案8：七夕节是中国民间的传统节日，为传统意义上的七姐诞。七夕节既是拜七姐节日也是爱情节日，又是一个以“牛郎织女”民间传说为载体，以祈福、乞巧、爱情为主题，以女性为主体的综合性节日。七夕节的传统习俗有拜七姐、祈福许愿、乞求巧艺、坐看牵牛织女星、祈祷姻缘、储七夕水等。七夕节这天，少女、妇女们三五成群约好了一起举行仪式，摆好瓜果祭品，沐浴斋戒，焚香礼拜。然后大家一起围坐在桌子前，一边吃瓜果点心，一边向织女默默许愿，如嫁个如意郎君、早生贵子之类。七夕节各项比赛中，穿针比赛其中最早开始流行的，大约从汉代开始。在七夕节这天，各家各户的女人女孩都会有出家门，大家聚在葡萄架下，一边猜女郎织女相会会说什么轻言密语，一边拿出自己的针线女红。通过穿针的方式乞巧，穿的最快者就是手最巧的。

答案9：七夕节是中国传统情人节，是情侣之间的节日。在七夕这一天，人们会祈求自己能够心灵手巧，同时也会许愿。此外，人们还会在晚上燃放鞭炮，吃糖果，换新被褥，摆放照片等。

答案10：七夕节是中国的传统节日，人们会进行许多习俗活动，如折桃枝、拜织女、揉木槿、吃巧果、洗梳具、染指甲、贺牛生日、拼七巧板、穿针乞巧、种生求子、供奉磨喝乐等。此外，七夕节也是中国的情人节，人们会穿针乞巧，祈祷福禄寿，礼拜织女，陈列花果、女红，各式家具、用具都精美小巧、惹人喜爱。
\end{CJK}
\tcblower
\begin{CJK}{UTF8}{gbsn}
\scriptsize
观点：七夕节是中国的传统节日，起源于牛郎织女的神话故事，是祈求姻缘、团圆和幸福的日子。

答案ID：答案1、答案2、答案3
观点：七夕节是中国的传统节日，人们会进行各种习俗活动，如拜织女、穿针乞巧等，祈求心灵手巧、获得美满姻缘。

答案ID：答案4、答案7、答案10
观点：七夕节是中国的传统情人节，是情侣之间的节日，人们会通过各种方式表达爱意。

答案ID：答案5、答案9
观点：七夕节是中国传统节日中最具浪漫色彩的一个节日，主要参与者是少女，节日活动以乞巧为主。

答案ID：答案6
观点：七夕节是中国民间的传统节日，为传统意义上的七姐诞，是一个以祈福、乞巧、爱情为主题，以女性为主体的综合性节日。

答案ID：答案8
\end{CJK}
\end{tcolorbox}

\clearpage

\begin{tcolorbox}[width=\textwidth]
\begin{CJK}{UTF8}{gbsn}
% \tiny
% \scriptsize
% \footnotesize
% \small
% \normalsize
% \large
% \Large
% \LARGE
% \huge
% \Huge
\scriptsize
Question: What is the Qixi Festival?

Answer 1: The Qixi Festival, also known as the "Cowherd and Weaver Girl Festival" or "Chinese Valentine's Day," is a traditional Chinese festival. It originated from the ancient myth of the Cowherd and Weaver Girl, where the Weaver Girl, said to be the seventh daughter of the Queen Mother of the West, fell in love with the Cowherd. However, they were separated by the Milky Way and could only meet once a year on the seventh day of the seventh lunar month. This festival is primarily associated with praying for love, reunion, and happiness. On Qixi Festival, people engage in various celebratory activities such as admiring lanterns, attending lantern festivals, paying respects to the Weaver Girl, and eating Qiao Guo (a type of pastry). Nowadays, Qixi Festival has also become a day for young people to express love and create a romantic atmosphere.

Answer 2: The Qixi Festival, also known as the "Cowherd and Weaver Girl Festival" or "Chinese Valentine's Day," is a traditional Chinese festival. It originated from ancient star worship, with the legend of the Cowherd (Altair, the Cowherd Star) and the Weaver Girl (Vega, the Weaver Star) meeting in the Milky Way. This story symbolizes people's yearning for and wishes for beautiful love. Qixi Festival is typically celebrated on the seventh day of the seventh lunar month, and activities on this day include praying for love, couples' reunions, and gift-giving. One of the most famous activities is "Bai Qijie," a ritual to pray for wisdom and craftsmanship from the Weaver Girl. Additionally, there are traditional customs like flying kites, stargazing, and enjoying lanterns.

Answer 3: The Qixi Festival, also known as the "Begging for Skills Festival" or "Daughter's Day," is a traditional Chinese festival rooted in ancient astronomical worship. Celebrated on the seventh day of the seventh lunar month, the day is believed to be when the Cowherd and Weaver Girl meet on the Magpie Bridge. On this day, people engage in various customs to pray for fulfilling love, such as praying to the Weaver Girl, begging for skills, paying respects to the Weaver Girl, and eating Qiao Guo. Simultaneously, Qixi Festival serves as China's Valentine's Day, where couples express love by exchanging gifts, having candlelit dinners, and more.

Answer 4: The Qixi Festival is a traditional Chinese festival where people participate in various customs and activities, including folding peach branches, worshipping the Weaver Girl, rubbing rose of Sharon flowers, eating Qiao Guo, washing and arranging tools, dyeing fingernails, celebrating the birthday of the Cow, playing Qijie puzzles, threading needles for skill-begging, sowing seeds for birth, offering "Mo He Le," and more. Additionally, Qixi Festival is also considered China's Valentine's Day, with customs like begging for skills, worshiping immortals, and threading needles to pray for a happy marriage.

Answer 5: The Qixi Festival is a traditional Chinese Valentine's Day, also known as "Daughter's Day" or "Qiaoqi Day," similar to Valentine's Day on February 14th. The seventh day of the seventh lunar month marks the meeting day of the Cowherd and Weaver Girl, a beautiful and eternal love story.

Answer 6: The Qixi Festival is the most romantic traditional Chinese festival, primarily for young girls. The festival focuses on the Begging for Skills ceremony, hence also known as "Begging for Skills Festival," "Girl's Festival," or "Daughter's Festival." On this evening, women engage in needle threading to pray for wisdom and craftsmanship, worshipping the Weaver Girl with sincere and solemn ceremonies. The festival is characterized by displaying flowers, fruits, and needlework, with various activities like making small dolls.

Answer 7: The Qixi Festival is a traditional Chinese festival where people perform various rituals and activities, such as needle threading for skills, spider-themed joyous activities, needle testing for skills, seed sowing for birth, offering "Mo He Le," worshiping the Weaver Girl, worshiping the Star of Marriage, sunning books, sunning clothes, celebrating the Cow's birthday, eating Qiao Guo, and more. Additionally, Qixi Festival features various small dolls displayed at home, with needle threading contests being one of the earliest popular activities, starting from the Han Dynasty. On Qixi Festival, women and girls gather under grape arbors, guessing what the Cowherd and Weaver Girl might say in secret while showcasing their needlework. The one who threads the needle the fastest is considered the most skillful.

Answer 8: The Qixi Festival is a traditional Chinese folk festival, celebrating the birth of the Seventh Sister in traditional beliefs. It is a festival of both worshiping the Seventh Sister and celebrating love, centered around the folk tale of the Cowherd and Weaver Girl. Traditional customs of Qixi Festival include worshipping the Seventh Sister, making wishes, begging for skills, observing the Cowherd and Weaver Girl stars, praying for marriage, storing Qixi water, and more. On Qixi Festival, groups of women and girls gather, set up fruit offerings, bathe and fast, burn incense, and pray to the Weaver Girl, making wishes like marrying a good husband and having children.

Answer 9: The Qixi Festival is a traditional Chinese Valentine's Day, a festival for couples. On this day, people pray for skillfulness and make wishes. Additionally, people light fireworks in the evening, eat sweets, change to new bedding, and display photos.

Answer 10: The Qixi Festival is a traditional Chinese festival with many customary activities, including folding peach branches, worshipping the Weaver Girl, rubbing rose of Sharon flowers, eating Qiao Guo, washing and arranging tools, dyeing fingernails, celebrating the Cow's birthday, piecing together Qijie puzzles, threading needles for skill-begging, sowing seeds for birth, offering "Mo He Le," and more. Moreover, the Qixi Festival is China's Valentine's Day, and people engage in activities such as threading needles to pray for skillfulness, offering prayers for good fortune and longevity, worshipping the Weaver Girl, displaying flowers and needlework, and using beautifully crafted furniture and utensils.

\end{CJK}
\tcblower
\begin{CJK}{UTF8}{gbsn}
\scriptsize
Viewpoint: The Qixi Festival is a traditional Chinese festival, originating from the myth of the Cowherd and Weaver Girl, a day for praying for love, reunion, and happiness.

Answer IDs: Answer 1, Answer 2, Answer 3
Viewpoint: The Qixi Festival is a traditional Chinese festival where people engage in various customs, such as worshiping the Weaver Girl and threading needles for skill-begging, praying for skillfulness, and achieving a happy marriage.

Answer IDs: Answer 4, Answer 7, Answer 10
Viewpoint: The Qixi Festival is China's traditional Valentine's Day, a festival for couples to express love in various ways.

Answer IDs: Answer 5, Answer 9
Viewpoint: The Qixi Festival is the most romantic traditional Chinese festival, primarily for young girls, with activities centered around begging for skills.

Answer ID: Answer 6
Viewpoint: The Qixi Festival is a traditional Chinese folk festival, celebrating the birth of the Seventh Sister in traditional beliefs. It is a comprehensive festival with themes of praying for good fortune, begging for skills, and love, with women as the main participants.

Answer ID: Answer 8
\end{CJK}
\end{tcolorbox}

\end{document}